\documentclass[letterpaper, 10 pt, conference]{ieeeconf}  

\IEEEoverridecommandlockouts                              

\usepackage[ruled,vlined,linesnumbered]{algorithm2e}
\usepackage{amsmath, graphicx}
\usepackage{balance}
\usepackage{amssymb,amsbsy}
\usepackage{breqn,bbm,xcolor}
\usepackage{multirow}
\usepackage[symbol,bottom]{footmisc}
\usepackage{soul}
\usepackage{url}
\usepackage{booktabs} 
\usepackage{tikz}
\usepackage[export]{adjustbox}
\usepackage{subcaption}
\usepackage{float}
\usepackage{tabularx}

\title{\LARGE \bf
CtRNet-X: Camera-to-Robot Pose Estimation in Real-world Conditions\\ Using a Single Camera
}

\author{Jingpei Lu$^{1,\dagger}$, Zekai Liang$^{1,\dagger}$, Tristin Xie$^{1}$, Florian Ritcher$^{1}$, Shan Lin$^{1}$, Sainan Liu$^{2}$, Michael C. Yip$^{1}$
\thanks{$\dagger$ These authors contributed equally.}
\thanks{$^{1}$Department of Electrical and Computer Engineering, University of California San Diego, La Jolla, CA 92093 USA.{\tt\small\{jil360, z9liang, tyx001, frichter, 
shl102, yip\}@ucsd.edu}}
\thanks{$^{2}$Intel Labs, USA. {\tt\small sainan.liu@intel.com}}
}

\begin{document}

\maketitle

\begin{abstract}
Camera-to-robot calibration is crucial for vision-based robot control and requires effort to make it accurate. Recent advancements in markerless pose estimation methods have eliminated the need for time-consuming physical setups for camera-to-robot calibration. While the existing markerless pose estimation methods have demonstrated impressive accuracy without the need for cumbersome setups, they rely on the assumption that all the robot joints are visible within the camera's field of view. However, in practice, robots usually move in and out of view, and some portion of the robot may stay out-of-frame during the whole manipulation task due to real-world constraints, leading to a lack of sufficient visual features and subsequent failure of these approaches. To address this challenge and enhance the applicability to vision-based robot control, we propose a novel framework capable of estimating the robot pose with partially visible robot manipulators. Our approach leverages the Vision-Language Models for fine-grained robot components detection, and integrates it into a keypoint-based pose estimation network, which enables more robust performance in varied operational conditions. 
The framework is evaluated on both public robot datasets and self-collected partial-view datasets to demonstrate our robustness and generalizability. As a result, this method is effective for robot pose estimation in a wider range of real-world manipulation scenarios.

\end{abstract}

\section{Introduction}

Estimating the Camera-to-Robot transform is crucial for manipulation, as it links the visual feedback from the camera to the space where the robot is operating, enabling accurate model-based robot arm manipulation with visual observations. 
Calibrating the Camera-to-Robot transform requires a significant amount of effort. Traditional calibration methods, such as \cite{garrido2014aruco, olson2011apriltag, fiala2005artag}, usually place fixed fiducial markers on the end-effector, collect images of several robot joint angles, and compute the transformation. These techniques have proved their advantage in generalizability and availability for different environments and robots.
However, such a procedure requires modification to the robotic system, which is not always possible, such as in instances where a dataset has already been collected \cite{padalkar2023open, khazatsky2024droid}.
Furthermore, the accuracy of the fiducial marker calibration approach is limited to the accuracy of the fiducial location relative to the robot.

The recent development of deep learning methods makes the markerless robot pose estimation possible, which can generally be divided into keypoint-based methods \cite{lu2022keypoint,lu2023markerless,lee2020camera} and rendering-based methods \cite{lu2022pose, hao2018vision, labbe2021single}. Contrary to classical approaches that need fiducial markers, deep learning-based pose estimation methods don't require cumbersome physical setups for calibration. Instead, they utilize deep neural networks for feature extraction or segmentation. The robot pose is then estimated using the keypoint features or the segmented robot masks.


\begin{figure}[t]
    \centerline{\includegraphics[width=\linewidth]{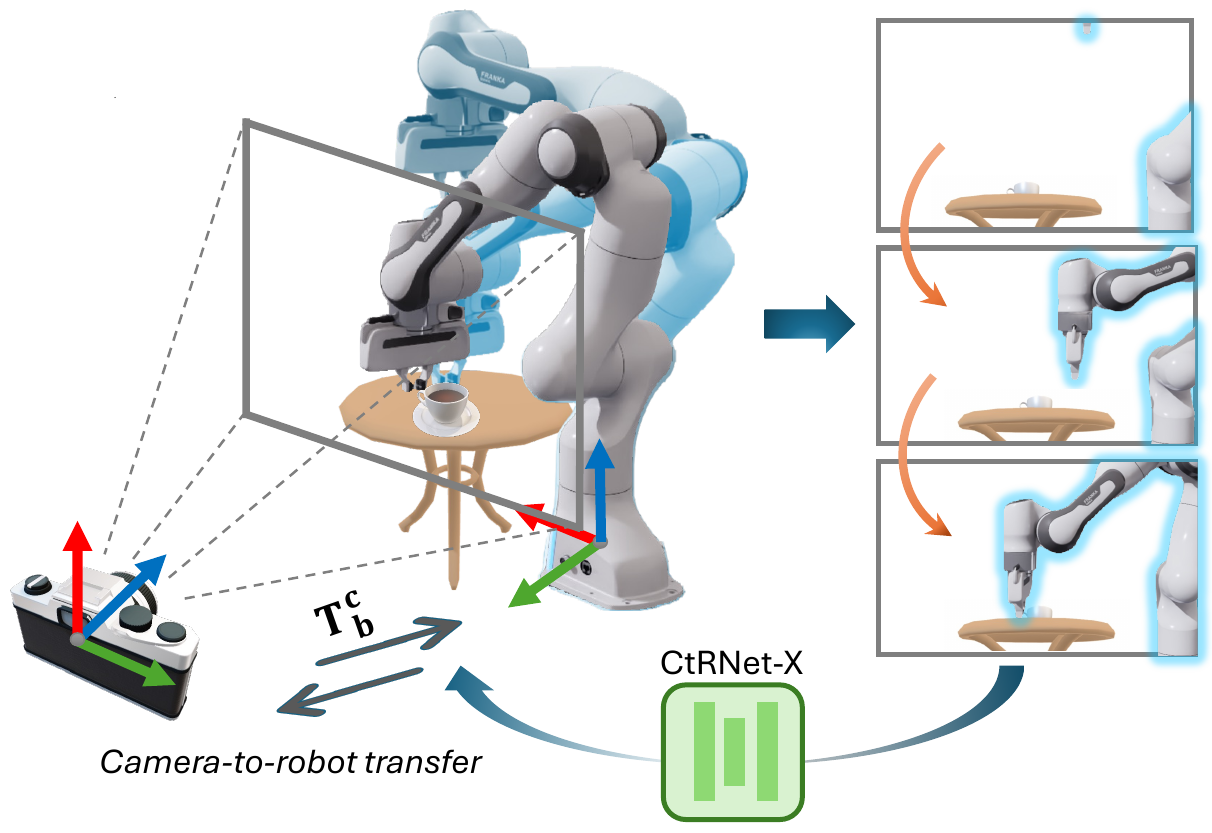}}
    \caption{In real-world robot manipulation scenarios, the camera does not always capture all the robot links, and the visibility of robot links changes from time to time. Our method leverages the limited available visual features within the camera view and achieves state-of-the-art performance on robot pose estimation.}
    \label{fig:overview}
\end{figure}

\begin{figure}[t]
    \centerline{\includegraphics[width=\linewidth,clip=true,trim={2mm 2mm 2mm 2mm}]{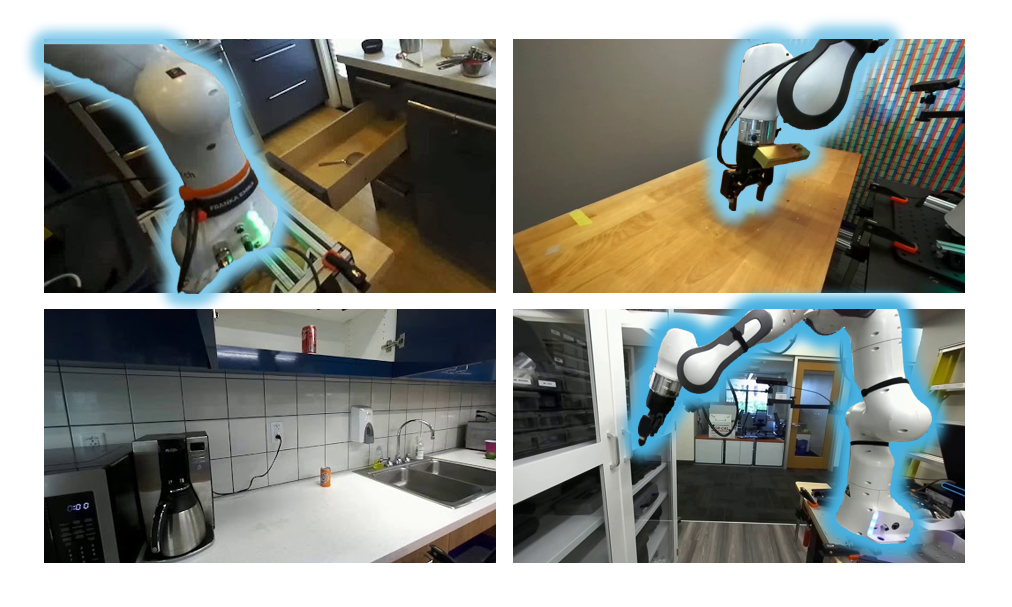}}
    \caption{Sample images are from DROID robot learning dataset \cite{khazatsky2024droid}. Often, only certain parts of the robot are visible in the camera view, and sometimes\ none of them are visible.}
    \label{classification example}
    \vspace{-0.14in}
\end{figure}

Despite considerable efforts to improve camera-to-robot pose estimation's flexibility, existing works have made a strong assumption that the entire robot arm is fully visible from the camera view. However, in real-world manipulation scenarios, the operating space is often limited, thus setting spatial constraints on camera placement \cite{levine2018learning, richter2021robotic, padalkar2023open, khazatsky2024droid}.
Additionally, the shape and size of target objects further complicate the trade-off between the capturing robot body and the manipulation targets. In such scenarios, the camera placement is typically driven more by the demands of the manipulation tasks instead of the need to observe all robot joints motion, as shown in Fig. \ref{fig:overview}. Consequently, video sequences with partially visible robot arm make up the majority of robotics manipulation datasets, such as Open X-Embodiment \cite{padalkar2023open} and DROID \cite{khazatsky2024droid}, as shown in Fig. \ref{classification example}. Existing methods often fail in situations where robots are only partially visible due to these real world constraints. Therefore, being able to estimate the robot pose with partial views is important from the practical aspect of robot manipulation scenarios where the camera can only capture a portion of the robot.


In this work, we introduce a novel framework for Camera-to-Robot Pose Estimation that extends the markerless robot pose estimation to partially visible scenarios. Our method integrates the Vision-Language Model (VLM) to detect the visible robot components and dynamically select the keypoints from visible robot links for the pose estimation. Moreover, we also improve keypoint detection performance by introducing the distribution-aware coordinate representation \cite{zhang2020distribution} to our previous development for the pose estimation network \cite{lu2023markerless}. We evaluate our framework on both fully visible and partially visible setups and achieve state-of-the-art performance on the public robot pose dataset and our self-collected dataset. 
In summary, our contributions are threefold:
\begin{itemize}
    \item We present a framework for markerless camera-to-robot pose estimation for more practical manipulation setups, where often only parts of the robot can be observed from a camera.
    \item We show the benefits of using the vision-language foundation model with few-shot learning for robot part detection and integrate it into the pose estimation framework.
    \item We show that our method achieves state-of-the-art performance compared to the existing methods on the benchmarking dataset while demonstrating the capability of estimating accurate camera extrinsic information for large-scale robot manipulation datasets.
\end{itemize}

\section{Previous Work}

\subsection{Camera-to-Robot Pose Estimation}

Traditionally, the camera-to-robot pose is calibrated using the fiducial markers \cite{garrido2014aruco,olson2011apriltag}. 
For articulated robots, the fiducial markers provide 2D point features on the robots, the 3D position of the markers can be calculated using robot kinematics, and the robot pose can be derived by solving a Perspective-n-Point problem~\cite{park_robot_1994,fassi2005hand,ilonen2011robust,horaud1995hand}.

As the field evolved, there was a shift towards markerless pose estimation. Initial efforts in this direction utilized depth cameras to localize articulated robots \cite{schmidt2014dart,pauwels2014real,michel2015pose,desingh2019factored}. With the rise of Deep Neural Networks (DNNs), a new paradigm emerged. DNNs, with their advantages of extracting point features without the need for markers, have significantly enhanced the performance of markerless pose estimation for articulated robots \cite{lambrecht2019towards,lee2020dream,lu2022keypoint,zuo2019craves}. Beyond keypoint-based methods, recent works \cite{labbe2021robopose,lu2023,chen2023easyhec} have demonstrated the potential of rendering-based methods. Benefiting from the dense correspondence provided by robot masks, rendering-based methods achieve state-of-the-art performance on robot pose estimation, but with compromise on the processing speed due to iterative render-and-compare. Most recently, \cite{lu2023markerless} proposed CtRNet, which uses robot masks to supervise the keypoint detector, achieving comparable performance to rendering-based methods while maintaining real-time inference speed.
Nonetheless, existing methods focus on scenarios where the robot manipulator is fully observable.
In real-world manipulations, it's non-trivial to set up the camera and the manipulator such that all the robot links stay within the camera view during the episodes, thereby diminishing the generalizability of existing methods when dealing with less constrained, real-world environments. In contrast, our proposed method overcomes this limitation by integrating a vision-language foundation model to detect the visibility of different robot parts and dynamically select the keypoints from visible robot parts for pose estimation.

\begin{figure*}[t]
    \centerline{\includegraphics[width=0.9\linewidth,clip=true,trim={0mm 0mm 0mm 0mm}]{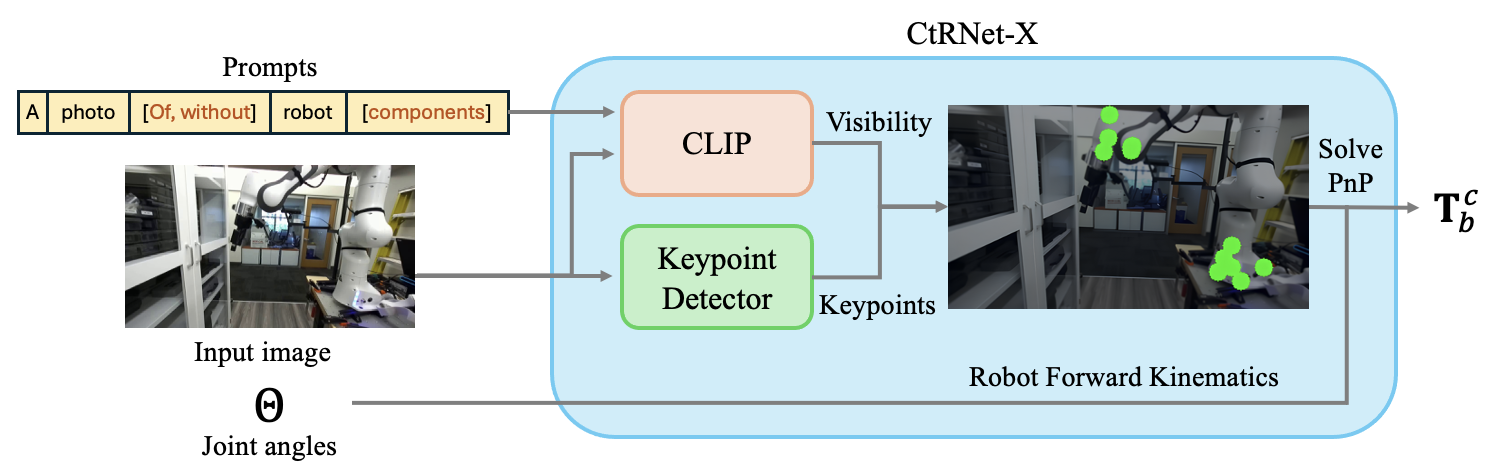}}
    \caption{Model inference pipeline. CtRNet-X estimates camera-to-robot transform given the images and the corresponding joint angles. The framework uses a set of structured prompts and the fine-tuned CLIP model to detect which robot parts are visible and dynamically adjusts the keypoint selection. The keypoint detector outputs 2D keypoints, and the corresponding 3D keypoints are obtained from the robot forward kinematics. Finally, a PnP solver is utilized to estimate the camera-to-robot transformation matrix given the selected keypoint correspondence. }
    \label{Method overview}
    \vspace{-0.14in}

\end{figure*}

\subsection{Robot Part Detection}


With the rapid development of deep learning, convolutional neural networks (CNNs) have demonstrated superior performance in object detection. The introduction of residual connections in ResNet \cite{he2016deep} makes it easier to construct deeper CNN architectures, hence achieving human-level performance on image recognition. However, deep neural networks typically require very large datasets to learn, which demands substantial hardware resources as well as labeling efforts. Furthermore, data is often not available due to not only the nature of the problem or privacy concerns but also the cost of data preparation \cite{parnami2022learning}.

Recent developments of Vision-Language Models (VLMs), such as CLIP \cite{radford2021learning}, have achieved remarkable performance on open-vocabulary object detection through training on large-scale datasets collected from the Internet. Recent research has focused on customizing the training and fine-tuning the CLIP model for specific downstream tasks. For example, CoOp \cite{zhou2022learning} optimizes the tokenized prompt vectors while freezing the model, reducing training time and maintaining the model's performance. Some following works \cite{bulat2023lasp}, \cite{chen2022plot}, \cite{lu2022prompt}, \cite{yao2023visual}, \cite{zhu2023prompt} further improve prompt learning's capability.  Additionally, Parameter-Efficient Fine-Tuning (PEFT) like Bitfit \cite{zaken2021bitfit} and Clip-adapter \cite{gao2024clip} focuses on VLM model optimization while attempting to minimize the number of training parameters, balancing training time and performance. However, the existing work mainly considers individual and common object classification, whereas their performance on fine-grained object parts detection is still unexplored. In this work, we explore several methods to tackle this challenge and demonstrate an effective way of fine-grained robot parts detection with few-shot learning samples.



\section{Methodology}
In this section, we introduce our framework for camera-to-robot pose estimation. The inference pipeline of our framework is shown in Fig. \ref{Method overview}. Our framework builds upon CtRNet \cite{lu2023markerless}, extending it to handle partially visible scenarios.
We utilize the vision-language foundation model, CLIP \cite{radford2021learning}, to identify the visible robot parts in the image frame hence selecting which keypoints to use for robot pose estimation.
Moreover, we also incorporate the Distribution-Aware coordinate Representation of Keypoint method (DARK \cite{zhang2020distribution}) into our framework to further enhance the performance of keypoint detection.
In Section \ref{sec:clip}, we detail our approach to fine-tune the CLIP model for robot part detection. In Section \ref{sec:ctrnet2.0}, we first provide an overview of CtRNet and then introduce our improvements.



\subsection{Few-shot Learning for Robot Parts Detection}
\label{sec:clip}

\textbf{VLM Fine-tuning}. Different from classifying the individual objects within images, our problem lies in detecting the components of the robot. Although vision-language foundation models demonstrate the capability of zero-shot object classification, they do not perform well in detecting robot parts. This is because the training data from the Internet lacks fine-grained semantic labels (e.g. robot end-effector, robot base).
In order to fine-tune the VLM to detect the robot parts in the images, we collected a small number of samples from the robot learning dataset, DROID \cite{khazatsky2024droid}, and investigated the few-shot transfer capability of the popular vision-language foundation model, CLIP \cite{radford2021learning}.

Training a large model like CLIP with a small dataset can be challenging. To address this, we employ the parameter-efficient fine-tuning method, Low-Rank Adaptation (LoRA \cite{hu2021lora}), and a strategic prompting method for fine-tuning the CLIP.
LoRA provides inspiration on freezing the pre-trained model weights and injects trainable rank decomposition matrices into each layer of the Transformer architecture. We add the LoRA module on both text and image encoders of the CLIP. For a forward linear passes $h = W_0 x$, we apply LoRA such that
\begin{equation}
h = W_0 x + \Delta W x = W_0 x + B A x
\end{equation}
where the $W_0$ is the pre-trained weight matrix of the self-attention module of CLIP, and $\Delta W$ is the trainable weight matrix. Specifically, $W_0 \in \mathbb{R}^{d \times k}$, $B \in \mathbb{R}^{d \times r}$ and $A \in \mathbb{R}^{r \times k}$, where the $r$ represents the low intrinsic dimension and $r \ll \min(d, k)$.
During the training, we only optimize the weights of these low-rank matrices $B$ and $A$ which have significant fewer parameters, while freezing the pre-trained weight. This approach allows for fast and efficient fine-tuning. 

\textbf{Prompt Strategy}. The prompt plays a crucial role in VLMs as a good prompt strategy can improve the performance of the model without extra effort. A typical prompt for CLIP is formulated as \texttt{A photo of \{object\}}. For classification, the \texttt{\{object\}} is replaced with different class labels, and the CLIP will select the class based on the cosine similarity between the image embedding and text embedding. 
In our scenario, the \texttt{\{object\}} is defined as the name of the robot components (e.g. robot base, robot end-effector). However, since the text of different robot parts is semantically similar, we have found that it is more effective to separate the queries for different components. Specifically, for each robot part, we input a pair of prompts (\texttt{A photo of robot \{component\}}, \texttt{A photo without robot \{component\}}). CLIP conducts binary classification for each component individually, thereby eliminating ambiguity when choosing from semantically close text embeddings.

To investigate the few-shot capability of VLM for robot part detection, we conducted a comparison of different few-shot learning approaches. These included parameter-efficient fine-tuning method (LoRA \cite{hu2021lora}), prompt learning method (CoOp \cite{zhou2022learning}), and traditional image classification using ResNet \cite{he2016deep} on the dataset we scraped from DROID. The quantitative results can be found in Section IV-B.

\subsection{Camera-to-Robot Pose Estimation}
\label{sec:ctrnet2.0}

\textbf{CtRNet Overview}. The Camera-to-Robot Pose Estimation Network (CtRNet \cite{lu2023markerless}) is the pioneer method for end-to-end robot pose estimation. The CtRNet includes a segmentation network, a keypoint detection network, and a differentiable Perspective-n-Point solver (BPnP \cite{chen2020end}). During the inference time, given the image frames and corresponding joint angles, the keypoint detector predicts the 2D keypoint coordinates on the robot manipulator, and the PnP solver estimates the robot pose with 2D-3D keypoint associations. The CtRNet is pre-trained on the synthetic dataset with ground-truth labels of segmentation masks and keypoint coordinates and is fine-tuned in the real-world data without labels in a self-training manner. In the self-training phase, a differentiable renderer is utilized to compute the robot mask based on pose estimation, and the masks obtained from the segmentation network are leveraged to provide image-level supervision to optimize the keypoint detector.

\textbf{Model Training}. In this work, we follow the training strategy of the CtRNet with modified keypoint placement. The CtRNet defines the keypoint at the location of each robot joint. To ensure the framework has a sufficient number of keypoint to estimate the pose for each image frame with partial view, we place $N$ ($N \geq 4$) number of keypoints for each robot link.
During the pre-training phase, CtRNet uses coordinate regression for training the keypoint detector, which minimizes the $L_2$ distance between the ground-truth and predicted keypoint coordinates.
Inspired by DARK \cite{zhang2020distribution}, we adapt heatmap regression for training the keypoint detector. The heatmap provides spatial support around the ground-truth location, taking into account contextual clues and the inherent ambiguity of the target position. Importantly, this approach can effectively reduce the risk of overfitting in the model during training, similar to the concept of class label smoothing regularization.
The heatmap regression minimizes the per-pixel $L_2$ distance between the predicted and ground-truth heatmap. In this work, we assume the heatmap should follow the Gaussian distribution. To supervise the heatmap prediction, the ground truth keypoint coordinates are encoded into the Gaussian heatmap, $\mathcal{D}$, as 
\begin{equation}
    \small
    \mathcal{D}(u, v) = \frac{1}{2\pi\sigma^2} \exp\left(-\frac{(u - u^*)^2 + (v - v^*)^2}{2\sigma^2}\right)
\end{equation}
where $u, v$ are pixel coordinates in the heatmap, $u^*, v^*$ are the ground truth keypoint coordinates, and $\sigma$ is the predefined spatial variance.
After pre-training on the synthetic dataset, we conduct self-supervised training on the real-world data without labels. The objective of self-training is to optimize the neural network parameters by minimizing the difference between the segmentation robot mask and the robot mask rendered based on the predicted pose. We follow the same self-training strategy as the CtRNet and the details can be found in \cite{lu2023markerless}. 

\textbf{Model Inferencing}. During the inference phase, we integrate the distribution-aware coordinate decoding \cite{zhang2020distribution} to extract 2D coordinates of keypoints from the predicted heatmap. We assume the predicted heatmap follows a 2D Gaussian distribution:
\begin{equation}
    \small
    \mathcal{G}(\mathbf{x}; \boldsymbol{\mu}, \Sigma) = \frac{1}{(2\pi) |\Sigma|^{1/2}} \exp\left(-\frac{1}{2}(\boldsymbol{x} -  \boldsymbol{\mu})^\top \Sigma^{-1} (\boldsymbol{x} - \boldsymbol{\mu})\right)
\end{equation}
where $\mathbf{x}$ is the pixel location, $\boldsymbol{\mu}$ is the Gaussian mean and $\Sigma$ is the distribution's covariance. In order to estimate $\boldsymbol{\mu}$, we follow the maximum likelihood estimation principle and transform the distribution function to the Log-likelihood function:
\begin{multline}
    \mathcal{P}(\boldsymbol{x}; \boldsymbol{\mu}, \Sigma) = \ln(\mathcal{G}) \\ 
    = -\ln(2\pi) - \frac{1}{2} \ln(|\Sigma|) - \frac{1}{2} (\boldsymbol{x} - \boldsymbol{\mu})^\top \Sigma^{-1} (\boldsymbol{x} - \boldsymbol{\mu})
\end{multline}
Following \cite{zhang2020distribution}, we can approximate $\boldsymbol{\mu}$ using the maximum activation $\mathbf{m}$ of the predicted heatmap, as the maximum activation generally represents a good coarse prediction that approaches $\boldsymbol{\mu}$. Then, we can approximate $\mathcal{P}(\boldsymbol{\mu})$ using the Taylor
series expansion evaluated at $\mathbf{m}$:
\begin{multline}
    \mathcal{P}(\boldsymbol{\mu}) = \mathcal{P}(\boldsymbol{m}) + \mathcal{D}'(\boldsymbol{m})(\boldsymbol{\mu} - \boldsymbol{m}) 
    \\ + \frac{1}{2}(\boldsymbol{\mu} - \boldsymbol{m})^\top \mathcal{D}''(\boldsymbol{m})(\boldsymbol{\mu} - \boldsymbol{m})
\end{multline}
Solving the above equation, we can obtain the equation to estimate $\boldsymbol{\mu}$ as 
\begin{equation}
      \boldsymbol{\mu} = \boldsymbol{m} - \left(\mathcal{D}''(\boldsymbol{m})\right)^{-1} \mathcal{D}'(\boldsymbol{m})
\end{equation}
where the \(\mathcal{D}''(\boldsymbol{m})\) and \(\mathcal{D}'(\boldsymbol{m})\) are the first and second image derivative of the predicted heatmap at the maximum activation $\mathbf{m}$. The detailed derivation can be found in \cite{zhang2020distribution}.

For robot manipulation tasks, the robot and camera positions often remain fixed throughout the episode. We can leverage this temporal consistency to improve the accuracy of estimation. Instead of estimating the robot pose for each single frame, we estimate it based on a batch of image frames from a single episode.
To perform batch estimation, we first use CLIP to predict which parts of the robot are visible in each image frame and select the keypoints that appear on the visible parts. Since we have more than enough keypoints to solve for the pose, we prioritize the keypoints with high confidence. We evaluate each keypoint based on the maximum activation value in the heatmap, denoted as $|\mathbf{m}|$, and filter out the keypoints with a maximum activation value below a certain threshold.
Finally, we combine all the reliable keypoints from multiple frames and use the PnP solver to estimate the robot pose.

\section{Experiments and Results}
\label{section:experiment}

\begin{table*}[ht]
    \centering
    
    \caption{Comparsion of top-1 accuracy (\%) and training time (s) of different fine-tuning methods on robot parts detection with few-shot learning. We evaluate the detection accuracy for the robot base (Base) and robot end-effector (EE). The results are based on the average of 3 seeds. Fine-tuning CLIP using LoRA achieves good performance with short training time.}
    \begin{adjustbox}{width=\textwidth,center}
    \begin{tabular}{lccccccccccccccc}
    
    \toprule
    \multirow{2}{*}{Method} & \multicolumn{3}{c}{0 shots} & \multicolumn{3}{c}{4 shots} & \multicolumn{3}{c}{8 shots} & \multicolumn{3}{c}{16 shots} & \multicolumn{3}{c}{32 shots} \\
    \cmidrule(lr){2-4} \cmidrule(lr){5-7} \cmidrule(lr){8-10}\cmidrule(lr){11-13}\cmidrule(lr){14-16}
     & EE (\%) & Base (\%) & Time (s) & EE  & Base  & Time & EE  & Base  & Time & EE  & Base  & Time & EE  & Base  & Time \\
    \midrule
    ResNet50 \cite{he2016deep} & - & - & - & 62.76 & 62.23 & \textbf{31.17} & 70.56 & 69.43 & \textbf{46.14} & 79.43 & 79.46 & \textbf{72.90} & 91.13 & 77.23 & 138.91 \\
    ResNet152 \cite{he2016deep} & - & - & - & 55.00 & 61.10 & 70.23 & 62.76 & 61.66 & 85.16 & 71.66 & 67.23 & 121.90 & 91.66 & 75.56 & 203.34 \\
    CoOp \cite{zhou2022learning}& 75.00 & 63.33 & - & \textbf{82.76} & 67.76 & 69.70 & \textbf{86.70} & 73.33 & 87.10 & 87.23 & 72.23 & 138.82 & 93.33 & 68.90 & 159.79 \\
    CLIP (Full Fine-tuning) & 75.00 & 63.33 & - & 76.66 & 72.20 & 316.06 & 82.23 & 72.76 & 330.69 & 86.13 & 75.00 & 365.27 & 90.00 & 80.00 & 450.81 \\
    CLIP (LoRA \cite{hu2021lora})& 75.00 & 63.33 & - & 76.66 & \textbf{81.13} & 40.70 & 81.10 & \textbf{80.56} & 57.89 & \textbf{92.76} & \textbf{80.56} & 89.51 & \textbf{96.70} & \textbf{87.23} & \textbf{108.33} \\
    \bottomrule
    \end{tabular}
    
    \end{adjustbox}
    \label{few shot}
\end{table*}

\begin{figure*}[ht]
    \centerline{\includegraphics[width=\linewidth,clip=true,trim={2mm 2mm 2mm 2mm}]{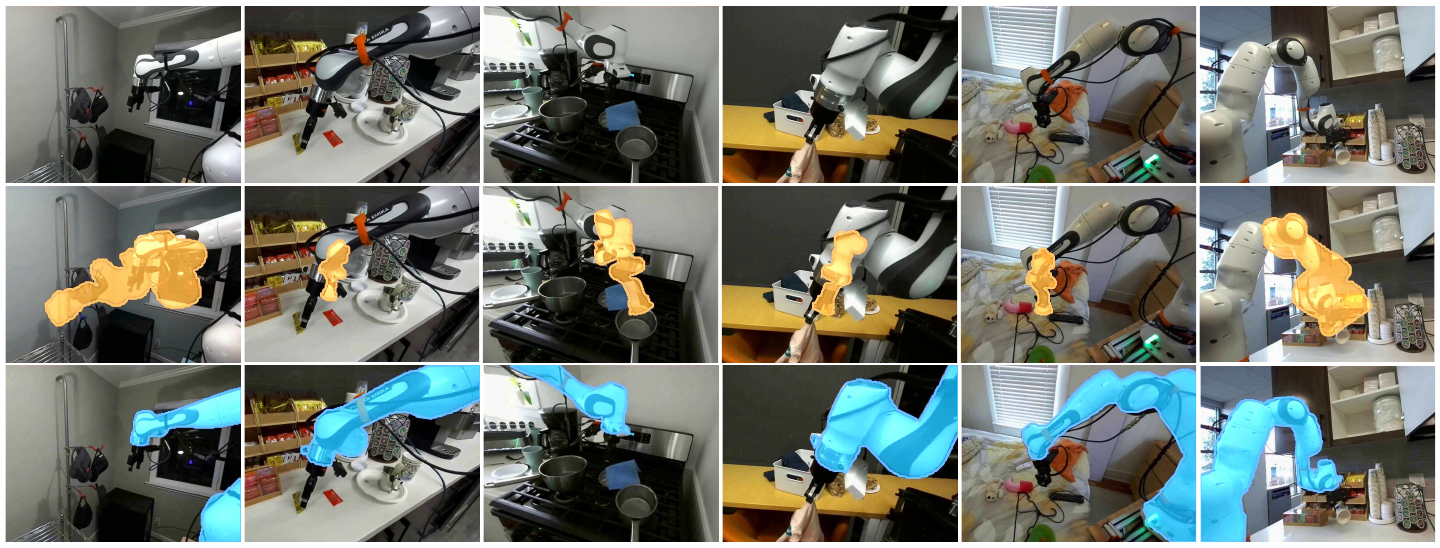}}
    \caption{Qualitative results of our method on the real-world manipulation dataset DROID \cite{khazatsky2024droid}. The first row is the raw image frames, the second is the robot masks rendered based on the estimation of the original CtRNet (orange), and the third row is the robot masks rendered based on the estimation of the CtRNet-X (blue).
    As shown above, CtRNet fails under real-world conditions whereas our method exhibits greater generalizability.}
    \label{qualitative_droid}
    \vspace{-0.1in}    
\end{figure*}

    \begin{table}[ht]
    \centering
    \caption{Performance comparison of different methods on DREAM-real dataset. We report the overall keypoint accuracy for the mean ADD and AUC of ADD.}
    \begin{tabular}{lccc}
    \toprule
    Method & Category & AUC $\uparrow$ &  Mean (m) $\downarrow$  \\
    \midrule
    DREAM-F~\cite{lee2020dream}  & Keypoint & 60.740 & 113.029 \\
    DREAM-Q~\cite{lee2020dream}  & Keypoint & 56.988 & 59.284 \\
    DREAM-H~\cite{lee2020dream} & Keypoint & 68.584 & 17.477\\
    RoboPose~\cite{labbe2021robopose} & Rendering & 80.094 & 0.020\\
    CtRNet~\cite{lu2023markerless} & Keypoint & 85.962 & 0.020\\
    CtRNet-X & Keypoint & \textbf{86.231} & \textbf{0.014}\\
    \bottomrule
    \end{tabular}
    \label{evaluation}
    \vspace{-0.14in}

    \end{table}

\subsection{Implementation Details}

We implemented the framework in Pytorch. The CLIP for robot parts detection is fine-tuned using an NVIDIA RTX 3090 GPU, while the pre-training and self-training of the keypoint detector are conducted on an NVIDIA RTX A6000 GPU.
For few-shot learning comparisons, the network or prompt parameters are trained for 200 epochs, given the small sample size and the models converge quickly. 
For fine-tuning the CLIP using LoRA, we apply low-rank matrices on the query, key and value matrices for both image and text encoders with \(r\) = 2.

For the keypoint detector, we pre-train the neural network on synthetic data from the DREAM dataset \cite{lee2020camera} and follow the training parameters from the CtRNet \cite{lu2023markerless}. The standard deviation of the Gaussian heatmap is set to 6 pixels. In the robot manipulation scenarios, the most common visible components are the robot end-effector and robot base. Hence, we simplify our framework to specifically recognize the robot end-effector and robot base. We place 6 keypoints for each of the links representing the robot end-effector and robot base (see example in Fig. \ref{Method overview}).
For self-training, we utilize the DREAM-real dataset and our self-collected Panda manipulation dataset.

\subsection{Few-shot Learning for Robot Parts Detection}
    
To examine the performance of VLM for robot parts detection, we compare popular fine-tuning methods, including Low-Rank Adaption \cite{hu2021lora}, prompt learning \cite{zhou2022learning}, and full-fine-tuning. Moreover, we also include classical object recognition technique using ResNet \cite{he2016deep} with ImageNet pretrained weights. For simplicity, we conduct the experiment on detecting the robot end-effector and robot base links. We train the networks on the training dataset scraped from DROID \cite{khazatsky2024droid} and evaluate the performance on a test dataset that has various unseen environments. We experiment with 3 random seeds and report the average top-1 accuracy for each category. The results are shown in Table \ref{few shot}.
We found that increasing the number of learning samples improves performance in general. However, the improvement becomes marginal when the training dataset becomes larger. Fine-tuning the CLIP with LoRA achieves better performance overall and requires less training time when learning with 32 shots.


\subsection{Experiment on DREAM-real Dataset}
We benchmark the robot pose estimation performance of CtRNet-X on the DREAM-real \cite{lee2020dream} dataset. The DREAM-real dataset consists of real-world images of the Franka Emika Panda robot arm captured from three different camera setups with total of around 57k image frames. We evaluate our method, together with other state-of-the-art robot pose estimation methods, on a single-frame setup.
We adopt average distance (ADD) metric to evaluate the pose estimation accuracy,
\begin{equation}
    ADD = \frac{1}{n} \sum_{i=1}^{n} \left\| \tilde{\mathbf{T}}_b^c \mathbf{p}_i - \mathbf{T}_b^c \mathbf{p}_i \right\|_2
\end{equation}
where \(\tilde{\mathbf{T}}_b^c\) and \(\mathbf{T}_b^c\) stand for the ground truth and estimated pose respectively. The area-under-the-curve (AUC) value and mean ADD are reported in Table \ref{evaluation}. Benefiting from the advanced distribution-aware coordinate decoding method, CtRNet-X achieves higher AUC and lower mean errors compared to existing methods.

\begin{figure}[t]
    \centering
    \includegraphics[width=0.9\linewidth]{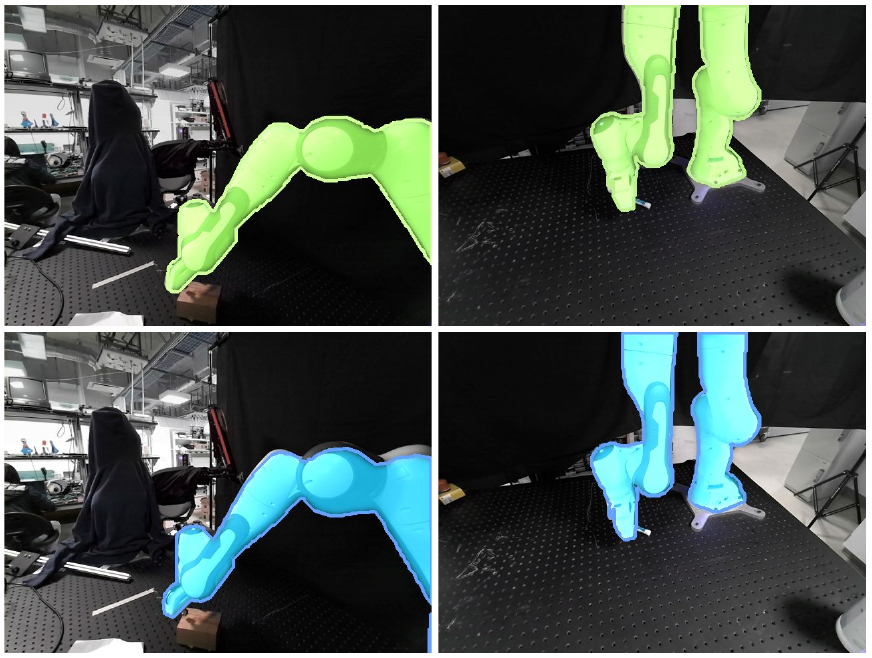}
    \caption{Qualitative results on Panda manipulation dataset. The first row is rendered robot masks using ground-truth extrinsic calibration (green) and the second row is the rendered robot masks using the pose from CtRNet-X (blue). }
    \label{fig:ours render}
    \vspace{-0.14in}
\end{figure}

\subsection{Experiment on Panda Manipulation Dataset}

The DREAM-real dataset only includes images where the robot arm is fully visible. To better evaluate our performance in real-world robot manipulation scenarios, we collected the Panda manipulation dataset. This dataset contains scenarios where only certain parts of the robot are visible during manipulation, and the robot arm is sometimes in and out of the image frame.

This dataset includes 60 video episodes: 30 robot-in-view episodes, where the visibility of the robot parts remains the same throughout the episode, and 30 robot-in-and-out episodes, where the visibility of the robot parts changes throughout the episode.
Each episode comes with synchronized robot joint angles and ground-truth camera-to-robot transformation. The camera-to-robot transformation is carefully calibrated using a checkerboard. To reduce calibration errors, we verify the calibration results by projecting the points at each link and overlaying the robot masks to ensure alignment with the images.
We compare our method with the original CtRNet \cite{lu2023markerless}, and report the ADD metric in Table \ref{our_ds_quantitative}. Additionally, we have provided the qualitative results in Fig. \ref{fig:ours render}. Our method demonstrates significant improvements in partial-view robot pose estimation compared to the previous method. By leveraging temporal consistency through batch estimation, CtRNet-X further improves accuracy by a significant margin.


\subsection{Experiment on DROID Dataset}

In this section, we demonstrate that our method can be used to obtain accurate extrinsic calibration for large robot learning datasets.
DROID \cite{khazatsky2024droid} is a large-scale in-the-wild robot manipulation dataset. We randomly sampled video episodes from the DROID and applied CtRNet-X to obtain the camera-to-robot transformation. We show some of the qualitative results in Fig. \ref{qualitative_droid}, where we render the robot masks based on the estimated robot pose.

To quantitatively evaluate our pose estimation performance on DROID, we use Segment Anything \cite{kirillov2023segment} to obtain the ground-truth robot masks and compute the Intersection over Union (IoU) for the rendered robot masks.
Due to labeling intensity for a hand-labeled ground-truth, we randomly selected 10 video episodes from DROID, which in total contain 3232 image frames, to label the ground-truth robot masks. The CtRNet-X achieves the average IoU of 0.8356.
We noticed that using the extrinsic information provided by the dataset, the average IoU of the rendered robot mask is 0.0186, demonstrating that the extrinsic calibration is prone to having errors which highlights the necessity for accurate extrinsic calibration using our method.


\begin{table}[t]
\centering
\caption{Qualitative results on Panda Manipulation Dataset. We report the overall mean and AUC of ADD with both single-frame and batch estimation.}
\scalebox{0.92}{
\begin{tabular}{lcccc}
\toprule
\multirow{2}{*}{Method} & \multicolumn{2}{c}{robot in view} & \multicolumn{2}{c}{robot in-and-out}\\
\cmidrule(lr){2-3} \cmidrule(lr){4-5}
 &  AUC $\uparrow$ & Mean (m) $\downarrow$  &  AUC $\uparrow$ & Mean $\downarrow$  \\
\midrule
CtRNet (single frame) &  16.764 & 0.381  &  35.944 & 0.335 \\
CtRNet-X  (single frame)& 60.317 & 0.059 & 59.828 & 0.056 \\
CtRNet-X  (batch)& \textbf{70.817} & \textbf{0.038} & \textbf{79.665} & \textbf{0.022} \\
\bottomrule
\end{tabular}}
\vspace{-0.14in}
\label{our_ds_quantitative}
\end{table}


\section{Conclusion}
    

We propose CtRNet-X, an end-to-end image-based robot pose estimation framework that can generalize to real-world robot manipulation scenarios. We employ the VLM, CLIP, for robot parts detection and dynamically select the keypoints of the visible robot parts. The robot pose is then estimated using a PnP solver with selected 2D and 3D keypoint correspondence.
We evaluate our method on the public robot pose dataset and self-collected manipulation dataset, demonstrating the superiority of our method in both fully and partially visible scenarios. 
Admittedly, the performance of the framework would be limited by the visible robot parts since we only utilize the robot end-effector and robot base links for pose estimation. However, our approach can be extended to finer granularity by including more robot parts. In the future, we will extend our method to different robots, incorporate kinematic uncertainty \cite{richter2021robotic}, and investigate the performance in more complex environments.

\balance
\bibliographystyle{ieeetr}
\bibliography{references}

\end{document}